\documentclass{article}
\usepackage{ijcai21}

\usepackage{times}
\usepackage{soul}
\usepackage{url}
\usepackage[hidelinks]{hyperref}
\usepackage[utf8]{inputenc}
\usepackage[small]{caption}
\usepackage{graphicx}
\usepackage{amsmath}
\usepackage{amsthm}
\usepackage{booktabs}
\usepackage{algorithm}
\usepackage{algorithmic}
\title{Exploiting Network Structures to Improve Semantic Representation for the Financial Domain}

\author{
Chao Feng$^1$
\and
Shi-jie Wei$^2$
\affiliations
$^1$University of Zurich\\
$^2$Harbin Institute of Technology\\
\emails
chao.feng2@uzh.ch,
2190400620@stu.hit.edu.cn
}

\begin{document}

\maketitle

\begin{abstract}
This paper presents the participation of the MiniTrue team in the FinSim-3 shared task on learning semantic similarities for the financial domain in English language. Our approach combines contextual embeddings learned by transformer-based language models with network structures embeddings extracted on external knowledge sources, to create more meaningful representations of financial domain entities and terms. For this, two BERT based language models and a knowledge graph embedding model are used. Besides, we propose a voting function to joint three basic models for the final inference. Experimental results show that the model with the knowledge graph embeddings has achieved a superior result than these models with only contextual embeddings. Nevertheless, we also observe that our voting function brings an extra benefit to the final system.
\end{abstract}

\section{Introduction}
Semantic representation and meaning representation have attracted the attention of linguists and philosophers for a long time~\cite{dowty2012word,chierchia2000meaning}. Human could share knowledge, experiences and thoughts through languages, i.e. words could be mapped to the concept space~\cite{vigliocco2007semantic}.  With the developing of Natural Language Processing technologies, words can be represented with dense vectors, which makes this mapping more promising~\cite{maarouf-etal-2020-finsim}. However, early studies have pointed that there is no one-to-one mapping between words meaning and concepts~\cite{murphy2004big,vigliocco2007semantic}.

An alternative is to map concepts to their hypernyms, which are the generic terms conceive the meaning of the concepts~\cite{maarouf-etal-2020-finsim}. FinSim-3 is such a shared task of hypernym categorization which focuses on financial domain. Financial entities have strong domain characteristics, which makes hypernym classification more difficult. It also means that this classification tightly coupled with semantic understanding. Therefore, we need a model which can fully mine semantic information as well as domain characteristics, to automatically classify financial concepts to their hypernyms.

Our work is done as part of the FinSim-3 (The 3rd Shared Task on Learning Semantic Similarities for the Financial Domain), which is a classification task for financial domain text data in English. We aim to combine sentence level embeddings learned by deep-learning models like BERT~\cite{devlin2018bert} with knowledge graph embedding model, to create a multi-label classification system. To solve the domain specific problem, we leverage various features of language models and knowledge graph embeddings model, and implement our system in four separate stages:
\begin{enumerate}
\item In order to solve the domain specific problem, we use two different pretrained language models in produce the sentence level embeddings, including the original English version BERT~\cite{devlin2018bert}, and the financial text pretrained language model FinBERT~\cite{araci2019finbert}.

\item To exploit the network structure of the concepts, we create a knowledge graph with external knowledge sources. Besides, a knowledge graph embedding algorithm has been applied to leverage the concepts structure information.

\item Three different models are built upon these different features and inference the output label independently.

\item We use a simple voting mechanism to integrate these three basic models, and to get the final label.
\end{enumerate}

The paper is organized as follows. We introduce the previous works related to our method in Section 2, and present the description of our method in Section 3. Then we group the experimental results in Section 4 before discussing the perspectives and concluding in Section 5.
\section{Related work}

As an important part of semantic relation extraction, hyponym-hypernym relationship identification has attracted the interest of NLP researchers for a long time. 

Traditionally, technologies of hyponym-hypernym detection make use of knowledge bases (e.g. WordNet, WikiData, etc.) to extract hypernymy relations~\cite{bordea-etal-2015-semeval}. However, recent approaches are more relied on data-driven methods. Text features (e.g. Character Count, Word Count, etc.) are extracted from the raw text. Then, a number of traditional machine learning algorithm could be applied for this problem \cite{saini-2020-anuj,anand-etal-2020-finsim20,bechara-etal-2015-miniexperts}.

Besides, approaches based on end-to-end neural networks offer an another promising alternative. Recurrent Neural Networks (RNN) and Long Short Term Memory (LSTM) networks are commonly used for Natural Language Processing tasks in earlier works. \cite{anand-etal-2020-finsim20} has applied LSTM models to hyponym-hypernym detection. But these approaches are mainly used the context-independent word embeddings or character embeddings. Since the pretrained transformer-based language models have been introduced to the NLP area, context-dependent embeddings are heavily used in different tasks. Researches have adopted BERT model to hyponym-hypernym detection task, and achieved highly scores~\cite{keswani-etal-2020-iitk-finsim}.

\section{System Architecture}
This section presents the architecture and the description of our system. Firstly, we show how to construct the knowledge graph from the provided extra knowledge sources, and how to train the knowledge graph embeddings. Secondly, three basic models are described in the Subsection 3.2. Lastly, we introduce the integration of our basic models.

\subsection{Knowledge Graph Embeddings}
As a powerful tool, knowledge graphs are usually used for knowledge representation. Within knowledge graphs, a factual triplet ($h$, $r$, $t$) represents a relation $r$ between a head-term $h$ and a tail-term $r$. Variety of NLP applications are based on knowledge graph technologies, such as question-answering system~\cite{hao-etal-2017-end} and recommendation systems~\cite{zhang2016collaborative}.

In this FinSim-3 task, the organizers provide a set of prospectuses to be used for embeddings training. On the other hand, we can also use these raw text data to construct a knowledge graph, which means if two-tuples of terms $h$ and $t$ are co-occurrence in one document $d$, we can group it as a triplet ($h$, $d$, $t$). As shown in Algorithm~\ref{alg:algorithm}, after traversing all the head and tail elements and articles, we can get a new knowledge graph $KG$ containing all co-occurrence relationships. 

\begin{algorithm}[H]
\caption{Construction of Knowledge Graph}
\label{alg:algorithm}
\textbf{Input}: Terms set $E$, documents set $D$.\\
\textbf{Output}:  A knowledge graph $KG$.\\
\begin{algorithmic}[1] 
\FOR{$h$ in $E$}
\FOR{$t$ in $E$}
\FOR{$d$ in $D$}
\IF {$d$ contains $h$ and $t$}
\STATE $KG$.append(($h$, $d$, $t$))
\ENDIF
\ENDFOR
\ENDFOR
\ENDFOR
\STATE \textbf{return} $KG$
\end{algorithmic}
\end{algorithm}

\begin{figure}[H]
\includegraphics[width=0.5\textwidth]{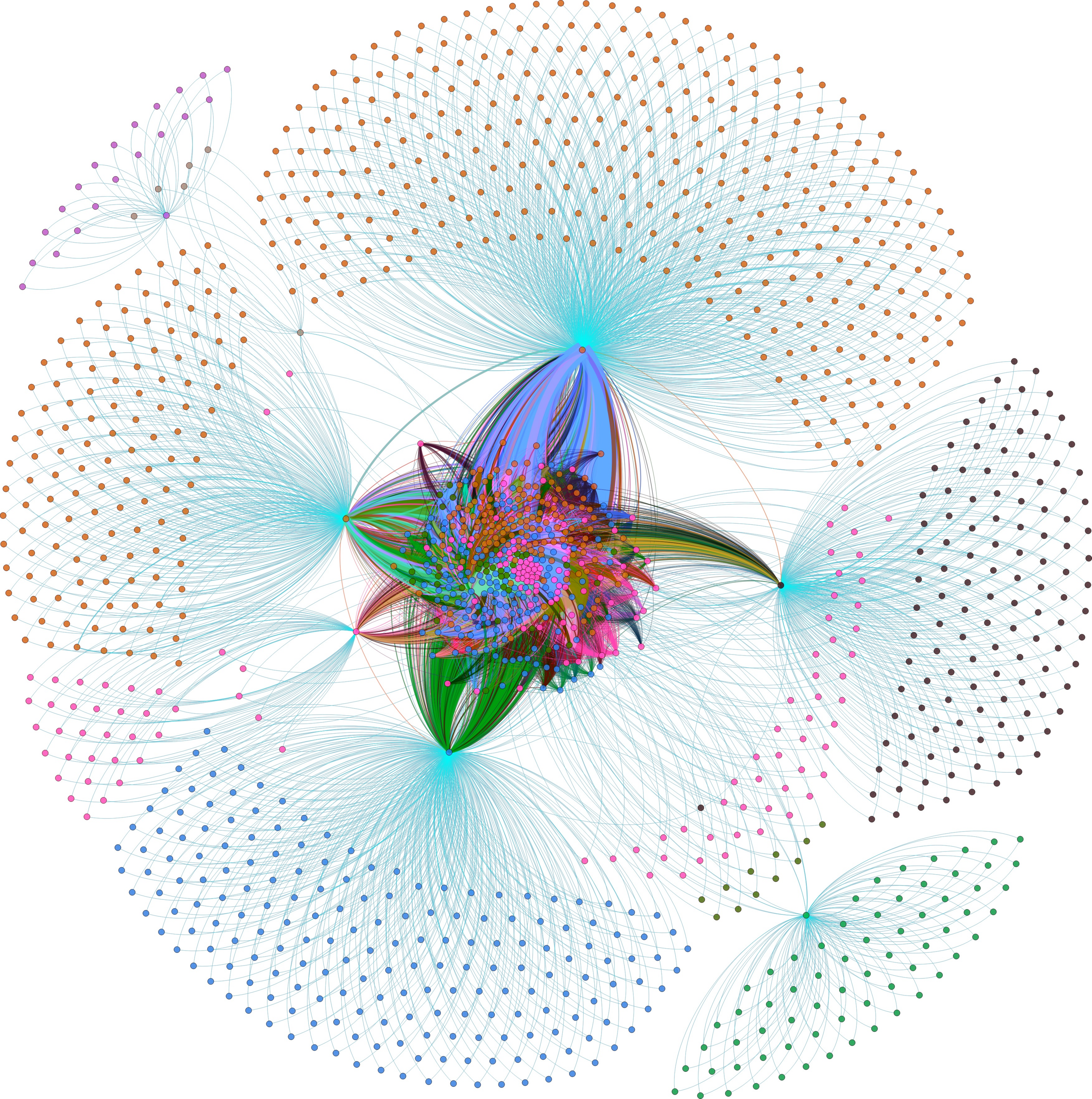}
\caption{The visualization of knowledge graph. Nodes means the terms, edges means the relations, and the colors present the different relations.} \label{kg}
\end{figure}

Figure \ref{kg} visualises the knowledge graph we created. The nodes present the terms of our training data-set. The edges present the co-occurrence relations between the head-term and the tail-term, and the colors of edges represent different documents.

Similar with word embeddings, using low dimensional vectors to represent entities and relationships has been proofed to be effective and efficient~\cite{bordes2013translating}. As we have created the knowledge graph $KG$, the next step is to train the knowledge graph embeddings. We use the RotatE algorithm to train our KG embeddings, which is proposed in~\cite{sun2019rotate}. Training with 1374 nodes and 1048575 edges, we get our 756-dimensions knowledge graph vectors.

\begin{figure}[tb]
\includegraphics[width=0.45\textwidth]{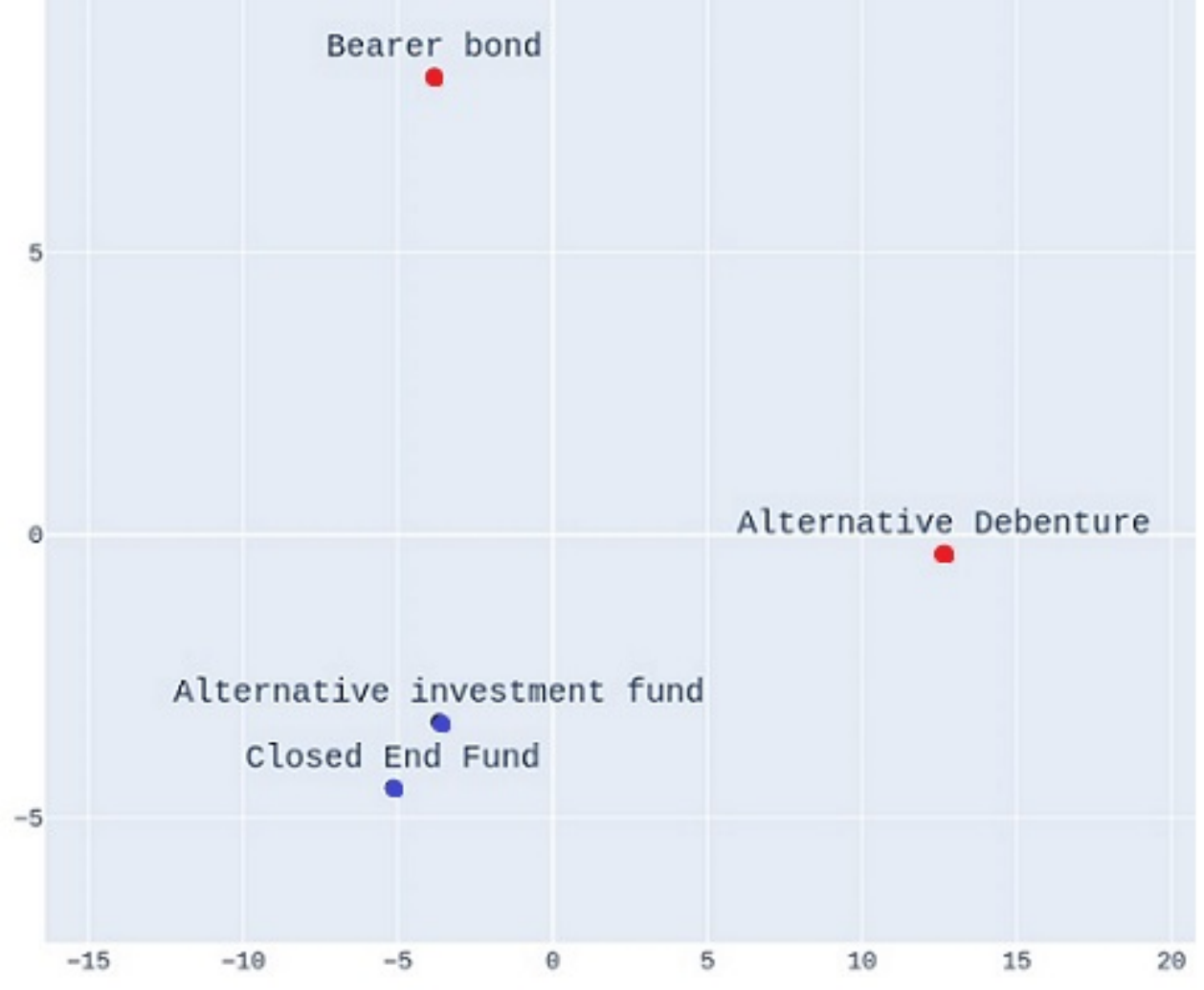}
\caption{A simple visualization of knowledge graph embeddings after PCA dimensionality reduction .} \label{kgb}
\end{figure}

Shown in Figure~\ref{kgb}, we use a Principal components analysis (PCA) method to cast these vectors into two-dimensions space, and randomly choose two categories of words to visualise these data, including Alternative Debenture and Bearer bond which belong to Bond category, and Alternative inverestment fund and closed End Fund which belong to Fund category. As we can see, there is a strong cohesion between terms within a same category. This means that our KG embeddings may provide additional information that can be used to help us classify effectively. 

\subsection{Basic Models}
We develop three independent basic models for label prediction by combining language models with simple neural networks. As before mentioned, two language models and one knowledge graph embeddings model are applied in our system, including BERT, FinBERT, and RotatE. 

\subsubsection{Basic Model One:}

This first basic model connects the language model, i.e. the BERT, with a feed forward network. After tokenization, the text data feeds into the language model, and generate a sentence embedding. After that, embedded data flows to a feed forward network for the final inference $I_1$.

\subsubsection{Basic Model Two:}

Research points out that the performance of original BERT model in domain related data is not as good as the domain specific language models~\cite{araci2019finbert}. On this count, we replace the BERT by financial version language model, FinBERT. Similar with the first model, sentence level embeddings flow into a inference network to get the final output $I_2$.

\subsubsection{Basic Model Three:}

To leverage the network structure information of the input data, we create a knowledge graph embedding model to obtain the network structure features from the extra knowledge source. In this time, we combine the features learned from FinBERT with the KG embeddings, and use a simple classifier network to get the final inference $I_3$.

\subsection{Voting Mechanism}
As before mentioned, $I_1$, $I_2$, $I_3$ are three inference values predicted by our basic models, and a simple voting mechanism is applied to count the final output. If two or more models draw the same inference label, this label will be our final prediction. The final output is predicted as:

\begin{equation}
f(x):=\left\{
\begin{aligned}
0&   \text{ if $I_1$+$I_2$+$I_3$ \textless  2 }\\
1&   \text{ if $I_1$+$I_2$+$I_3$ \textgreater= 2}
\end{aligned}
\right.
\end{equation}

\begin{figure}
\includegraphics[width=0.5\textwidth]{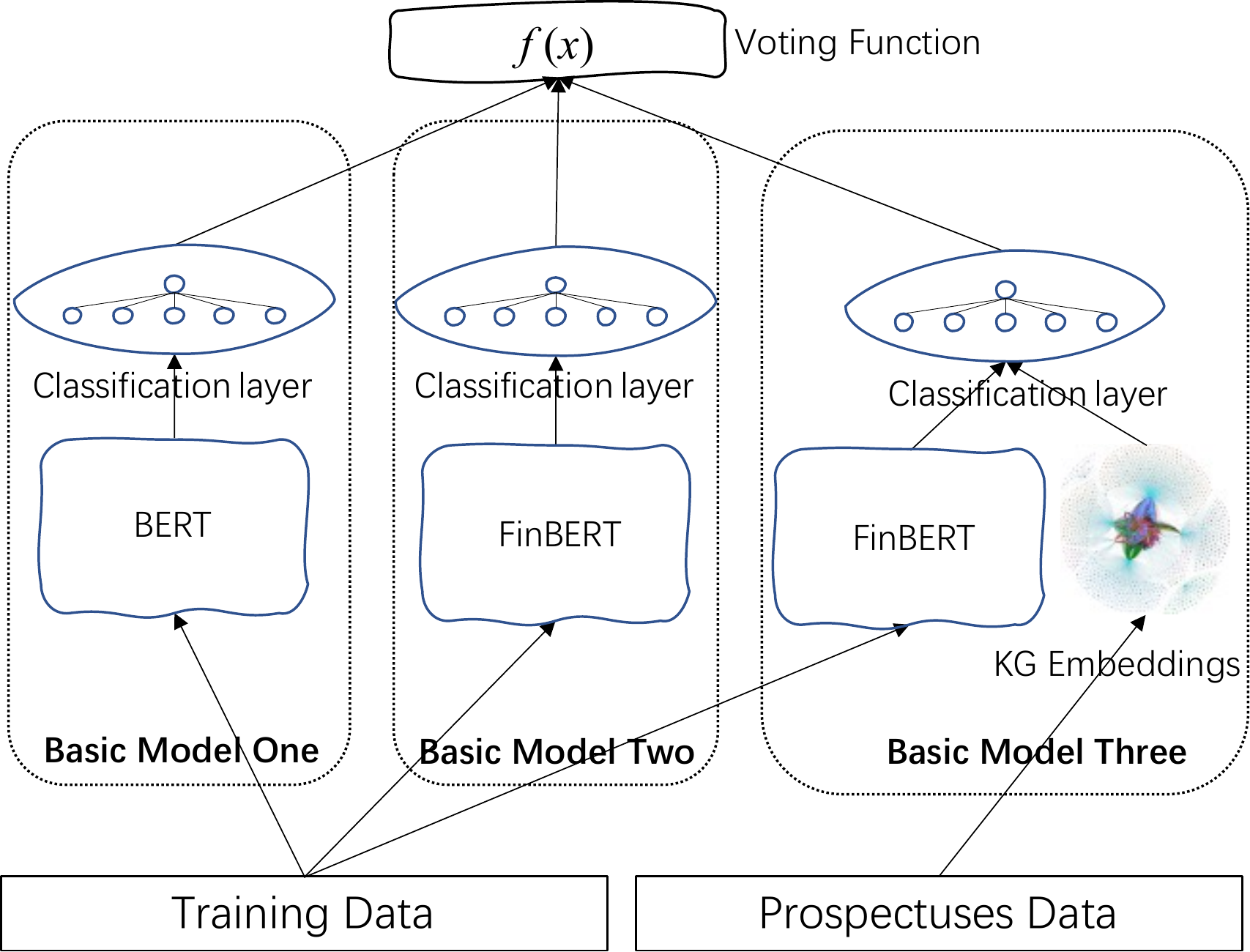}
\caption{The final model combines three basic models with a voting mechanism} \label{fig1}
\end{figure}

The architecture of our model is shown in Figure \ref{fig1}. As we can see, the input text date feeds into each basic model, and then flows to each inference network respectively. Finally, all of this information is combined in the voting function for the prediction.

\section{Experiments and Results}
In this section, we present the description of datasets, the setup of our experiments, and the final results of our experiments. Experimental validation is conducted on the FinSim-3 training and test corpus. The datasets are described in subsetion 4.1 followed by the analysis of our final results in subsetion 4.2.

\subsection{Data Description}
In this shared task, the organizers give a list of selected terms from the financial domain. Participants are required to design and implement a classifier to automatically classify these financial terms to the most relevant hypernym concepts. 

As shown in Table~\ref{tab:corpus}, the train and the test sets are not of equal size. There are 1050 terms in the train set, and 326 terms in the test set. 

\begin{table}[H]
\centering
\begin{tabular}{lll}
\hline
\textbf{Corpus}  & train & test \\
\hline
\textbf{Number of terms}  & 1050 & 326 \\
\hline
\end{tabular}
\caption{Dataset of terms for FinSim-3.}
\label{tab:corpus}
\end{table}

In this task, we need to classify these terms into categories of financial instruments, including 'Bonds', 'Forward', 'Funds', 'Future', etc. Table~\ref{tab:termdis} illustrates the distribution of the terms in each category. As we can see, the number of terms in each label are not balanced, there are huge gaps among these numbers of different categories. For example, about one-fourth of terms belong to the ’Equity Index' label, but only eight of them belong to 'Securities restrictions'. This data-imbalance problem makes the task more difficult.

\begin{table}
\centering
\begin{tabular}{lll}
\hline
Corpus  & Term category & Number of terms \\
\hline
& Bonds & 55\\
& Central Securities Depository & 107\\
& Credit Events & 18\\
& Credit Index & 129\\
& Debt pricing and yields & 58\\
& Equity Index & 286\\
& Forward & 9\\
& Funds & 22\\
Train & Future & 19\\
& MMIs & 17\\
& Option & 24\\
& Parametric schedules & 15\\
& Regulatory Agency & 205\\
& Securities restrictions & 8\\
& Stock Corporation & 25\\
& Stocks & 17\\
& Swap & 36\\
\hline
\end{tabular}
\caption{Number of terms in each financial instruments.}
\label{tab:termdis}
\end{table}

\subsection{Results and Discussion}
The evaluation of this hypernym identification task is mainly based on Accuracy score and Average Rank score. Our results are gathered in Table \ref{tab:trainres}, which contains the Accuracy and Average Rank scores for the basic models and the final systems. M1 - M3 refer to our basic model one - three, and M4 refers to our final model. As we can only submit three results to the organizers, so only the results of M2 - M4 are shown in the table.

\begin{table}[H]
\centering
\begin{tabular}{llll}
\hline
Corpus  & Model & Accuracy & Average Rank \\
\hline
& M1(BERT) & 0.944 & 1.168\\
train & M2(FinBERT) & 0.971 & 1.065\\
& M3(M2 + KG emb) & 0.972 & 1.068\\
& M4(M3 + Voting) & 0.973 & 1.051\\
\hline
& M2(FinBERT) & 0.825 & 1.337 \\
test & M3(M2 + KG emb) & 0.855 & 1.346\\
& M4(M3 + Voting) & 0.865 & 1.315\\
\hline
\end{tabular}
\caption{Evaluation Results: Accuracy and Average Rank.}
\label{tab:trainres}
\end{table}

\textit{\textbf{Q1.} Is the domain specific language model brings additional benefits to the system?}

The answer is simple yes. As we can see int Table~\ref{tab:trainres}, models with FinBERT are better than the original BERT model. With simply replace the BERT with FinBERT, our model gets a 3\% extra performance, which strongly support our hypothesis.

\textit{\textbf{Q2.} Is the knowledge graph embeddnigs brings additional benefits to the system?}

Back to the Table~\ref{tab:trainres}, our M3, FinBERT with KG embeddings, is better than the model with only FinBERT in both train and test sets. Especially in the test set, KG embeddings improve our model from 82.5\% to 85.5\%, which is a huge increase.

\textit{\textbf{Q3.} Is the voting function brings additional benefits to the system?}

\begin{figure}
\includegraphics[width=0.5\textwidth]{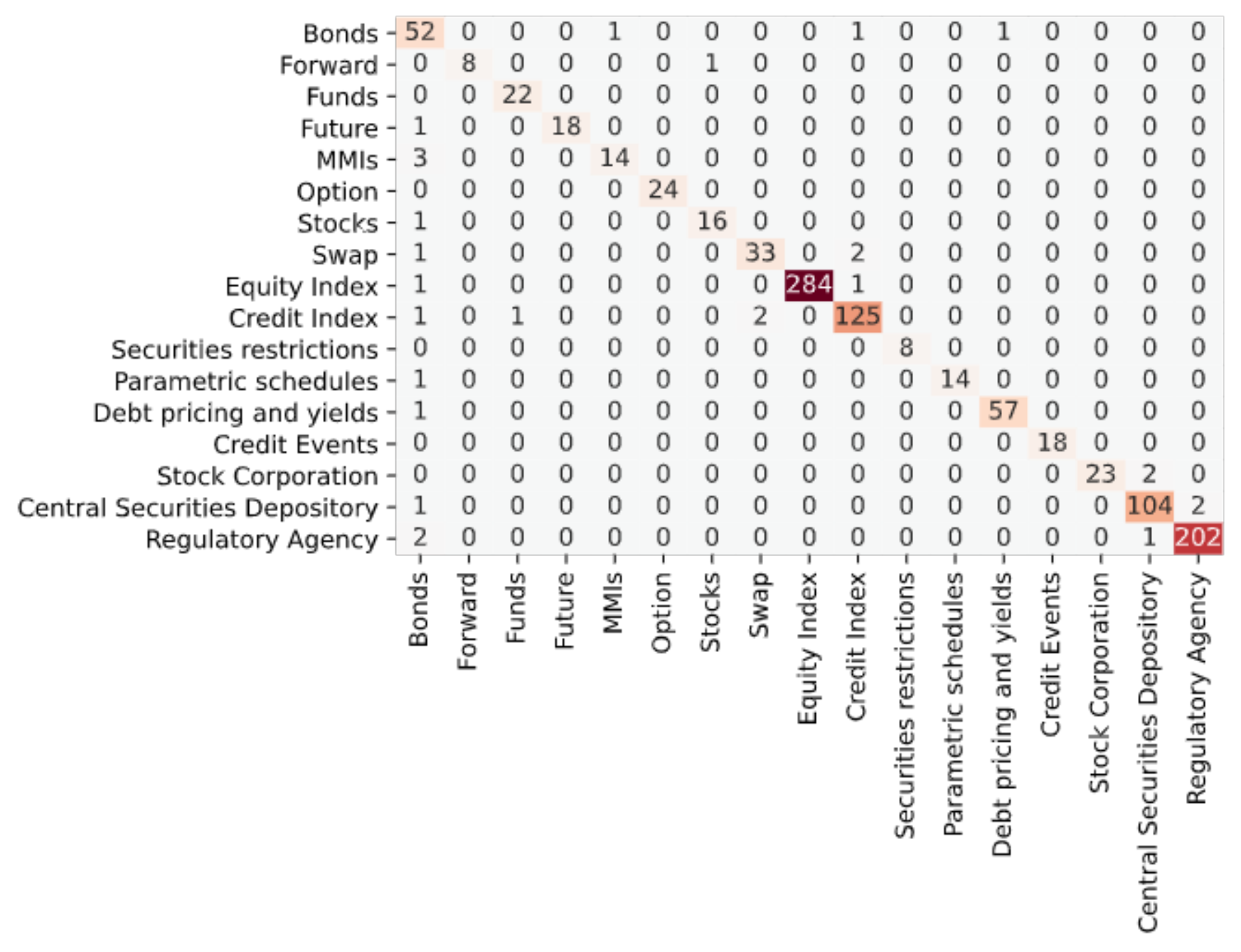}
\caption{Heat map for the confusion matrix of our final model} \label{fig2}
\end{figure}

As we can see, the our final system (with a simple voting mechanism) performs the best Accuracy and Average Rank scores among these four systems. We also care about where are the errors come from for our system. So we draw a confusion matrix of our final model in Figure~\ref{fig2}. As we can see, most of the mis-classification terms are incorrectly labeled as 'Bond'. The reason for this error may be the data imbalance problem, so that the weights of each label are unbalanced in our model.

\section{Conclusions}

In this paper, we implement a classification system with combining the language models and a knowledge graph embeddings model. Our system has been evaluated on the FinSim-3 shared task. The evaluation results has shown that our system is domain robust, and the KG embeddings and voting function indeed bring in extra benefits to our final system. Our experiments prove that the pre-trained language model is also highly adaptable to financial domain texts, and the network structure can highly leverage the predictive ability of the multi-label model system.

\bibliographystyle{named}
\bibliography{ijcai21}

\end{document}